\documentclass[sigconf, authorversion, screen]{acmart}

\usepackage{algpseudocode}
\usepackage[font=bf]{caption}
\usepackage{subcaption}

\AtBeginDocument{%
  \providecommand\BibTeX{{%
    \normalfont B\kern-0.5em{\scshape i\kern-0.25em b}\kern-0.8em\TeX}}}

\copyrightyear{2022} 
\acmYear{2022} 
\setcopyright{acmlicensed}\acmConference[FDG '22]{FDG '22: Proceedings of the 17th International Conference on the Foundations of Digital Games}{September 5--8, 2022}{Athens, Greece}
\acmBooktitle{FDG '22: Proceedings of the 17th International Conference on the Foundations of Digital Games (FDG '22), September 5--8, 2022, Athens, Greece}
\acmPrice{15.00}
\acmDOI{10.1145/3555858.3555896}
\acmISBN{978-1-4503-9795-7/22/09}

\acmConference[FDG '22]{Foundations of Digital Games 2022}{September 5--8,
  2022}{Athens, Greece}

\begin{document}

\title{EvolvingBehavior: Towards Co-Creative Evolution of Behavior Trees for Game NPCs}

\author{Nathan Partlan}
\email{partlan.n@northeastern.edu}
\orcid{0000-0002-9453-2117}
\affiliation{%
  \institution{Northeastern University}
  \streetaddress{360 Huntington Ave}
  \city{Boston}
  \state{MA}
  \country{USA}
  \postcode{02115-5000}
}

\author{Luis Soto}
\email{soto.lu@northeastern.edu}
\affiliation{%
  \institution{Northeastern University}
  \streetaddress{360 Huntington Ave}
  \city{Boston}
  \state{MA}
  \country{USA}
  \postcode{02115-5000}
}

\author{Jim Howe}
\email{howe.ha@northeastern.edu}
\affiliation{%
  \institution{Northeastern University}
  \streetaddress{360 Huntington Ave}
  \city{Boston}
  \state{MA}
  \country{USA}
  \postcode{02115-5000}
}

\author{Sarthak Shrivastava}
\email{shrivastava.sa@northeastern.edu}
\affiliation{%
  \institution{Northeastern University}
  \streetaddress{360 Huntington Ave}
  \city{Boston}
  \state{MA}
  \country{USA}
  \postcode{02115-5000}
}

\author{Magy Seif El-Nasr}
\email{mseifeln@ucsc.edu}
\orcid{0000-0002-7808-1686}
\affiliation{%
  \institution{University of California, Santa Cruz}
  \streetaddress{1156 High St}
  \city{Santa Cruz}
  \state{CA}
  \country{USA}
  \postcode{95064-1099}
}

\author{Stacy Marsella}
\email{s.marsella@northeastern.edu}
\orcid{0000-0002-5711-7934}
\affiliation{%
  \institution{Northeastern University}
  \streetaddress{360 Huntington Ave}
  \city{Boston}
  \state{MA}
  \country{USA}
  \postcode{02115-5000}
}

\renewcommand{\shortauthors}{Partlan, Soto, Howe, Shrivastava, Seif El-Nasr, and Marsella}

\begin{abstract}
To assist game developers in crafting game NPCs, we present EvolvingBehavior, a novel tool for genetic programming to evolve behavior trees in Unreal\textregistered Engine 4. In an initial evaluation, we compare evolved behavior to hand-crafted trees designed by our researchers, and to randomly-grown trees, in a 3D survival game. We find that EvolvingBehavior is capable of producing behavior approaching the designer's goals in this context. Finally, we discuss implications and future avenues of exploration for co-creative game AI design tools, as well as challenges and difficulties in behavior tree evolution.
\end{abstract}



\maketitle

\section{Introduction}

Game developers carefully craft non-player characters (NPCs) to support their game design goals, focusing on creating a specific player experience. But they may not always have specific training, technical knowledge, or access to dedicated support for implementing their desired NPC behavior. They may also find that players interact in unexpected ways with their games, and may want help developing NPCs that can handle unusual or emergent gameplay situations. Game developers, especially those working alone or with small teams and limited resources, may not have expertise in developing robust and effective game AI. Tools that could generate or suggest agent behavior might support designers in this work.

We should consider, then, whether it is possible to make an AI-enabled, computationally co-creative tool that could aid and collaborate with game developers in crafting and refining NPC behavior. As a prerequisite, however, such a tool must be able to produce NPC behavior to meet particular design goals. It must also work with existing game engines and game AI frameworks.

These problems can be considered a form of ``program synthesis,'' the automated creation of programs, an active area of research~\cite{gulwani_program_2017}. Synthesizing NPC behavior presents some unique challenges: it requires simulation to determine fitness (rather than having a direct mapping from input to desired output, as in many programs), and that evaluation must relate to the designer's goals for the behavior. It is important to investigate, then, how program synthesis methods can generate NPC behavior -- with the guidance, and meeting the goals, of a human designer -- within a reasonable amount of time.

To approach this problem, we choose to focus on behavior trees, a popular game AI design architecture~\cite{champandard_behavior_2013,isla_handling_2005}, in the professional and commonly-used Unreal\textregistered Engine 4 game engine\footnote{Unreal\textregistered is a trademark or registered trademark of Epic Games, Inc. in the United States of America and elsewhere.}. Specifically, we explore the combination of behavior trees and genetic programming, a tree-structured program synthesis technique that maps naturally to the behavior tree architecture and that has shown promise in creating behavior trees for simple 2D games~\cite{colledanchise_learning_2018,zhang_behavior_2018}.

In this work, we present EvolvingBehavior\footnote{EvolvingBehavior and its code are available at \url{https://evolvingbehavior.npc.codes}}, a novel tool that employs genetic programming to evolve modifications to handcrafted behavior trees as part of a co-creative game design process. As an initial evaluation of the tool, we describe an experiment that compares evolved behavior trees to trees hand-designed by our researchers and to randomly-grown trees for zombies in a 3D survival game, exploring the results through quantitative evaluation and in-depth discussion. We find that the evolved trees exhibit clearly stronger similarities to hand-designed trees than randomly-grown trees. Finally, we discuss implications for co-creative game AI design tools, challenges in automatic evolution of behavior trees, and potential future research to empower game designers to use these tools to achieve their NPC design visions.

\section{Related Work}

\subsection{Program Synthesis and Evolving Game AI}

We can think of generating AI behavior for agents in games as a form of program synthesis. Program synthesis is a hard problem, requiring search over a large space of possible program designs~\cite{gulwani_program_2017}. Stochastic techniques, such as genetic algorithms, are often helpful in navigating these very large decision spaces~\cite{gulwani_program_2017}, and can often generate relatively concise solutions.

Genetic programming comprises a class of evolutionary algorithms for program synthesis, operating on populations of tree-structured programs~\cite{koza_genetic_1992}. These algorithms measure the fitness of each population member by some specified metric(s), aiming to re-combine, modify, and improve the performance of population members over multiple generations. The name relates the technique vaguely to biological evolution, though many differences apply; i.e., selection is entirely artificial, mutation is intentional and simplistic, etc. We note ethical concerns with this comparison in Section~\ref{sec:ethics}.

Researchers have long sought to automatically evolve AI in games. Luke~\cite{luke_genetic_1998} applied genetic programming to evolve AI agents for a robot soccer competition. They used a custom-designed set of functions and operators, requiring a strongly-typed genetic programming approach. This was followed by similar efforts for other simple games~\cite{fogel_self-learning_2004,langdon_evolutionary_2005,alhejali_evolving_2010}, and more recently for high-level strategies in the popular RTS StarCraft~\cite{garcia-sanchez_towards_2015}. Some researchers evolved linear lists of instructions or rules~\cite{vowk_evolutionary_2004,priesterjahn_evolution_2007}. Most other early efforts were focused on parameter tuning of pre-defined behaviors, rather than generating behavior directly~\cite{bakkes_team_2004,cole_using_2004,soule_darwins_2017}. More recent attempts have focused mostly on neuroevolution, which evolves the structure and parameters of neural networks~\cite{stanley_evolving_2005,hastings_evolving_2009,risi_neuroevolution_2015,bulitko_-life_2019}, or, in one case, on combining reactive planning with search using an evolving evaluation function~\cite{neufeld_evolving_2019}. Because they used custom instruction sets, parameters, or neural networks designed for a specific game, it would require significant effort to translate these techniques to other games, and the results may not be human-interpretable.

\subsection{Behavior Trees}

Behavior trees are a popular architecture for game AI~\cite{champandard_behavior_2013}, stemming from the system developed by Isla for the Halo game series~\cite{isla_handling_2005}. They are tree-structured, making them suitable for visual editing. They provide clear hierarchy and control flow, with internal \emph{composite} nodes controlling execution, and leaf \emph{task} nodes to perform the selected behavior. For example, a ``Selector'' is a composite that runs each of its children in turn, stopping after the first successful child, and ``Move to location'' might be an example of a task. In many implementations, such as in Unreal Engine 4, \emph{decorator} nodes attach to other individual nodes and modify their execution or results. For example, a decorator might check for a particular condition, preventing execution of the child otherwise.

\subsubsection{Learning from Demonstration}

Some researchers have begun to explore learning behavior trees from demonstration~\cite{tomai_adapting_2014,robertson_building_2015,sagredo-olivenza_trained_2017}. This has the advantage of requiring less manual specification of objective functions than general search, because demonstrations serve to provide the objective. These methods tend to generate very complex trees that mimic each demonstration, however, and have yet to overcome the difficulty of simplifying the results to a compact and human-interpretable form. Research has shown that designers need detailed control over the final AI behaviors~\cite{partlan_design-driven_2021}, which requires the trees to be concise enough for hand-modification. 

\subsubsection{Generating Behavior Trees by Evolution}

We are not the first to try building behavior trees automatically using genetic algorithms, though most prior efforts have focused on generating behavior for specific, 2D games with simple and clear goals, rather than on creating a flexible tool for game designers to define their own design goals. Lim et al.~\cite{lim_evolving_2010} evolved small behavior trees to control specific strategic decisions, combining them manually to build an AI to play DEFCON, a strategy game. Perez et al.~\cite{perez_evolving_2011} evolved behavior trees to play Mario. Rather than evolving trees directly, they used grammatical evolution, constraining the structure to ``and-or'' trees. In continued research, they explored a potential advantage of using behavior trees in evolution: namely, choice over the level of granularity at which to represent conditions and actions~\cite{nicolau_evolutionary_2017}. Our work seeks to investigate whether a genetic-programming-based approach can enable evolution in varying 3D environments, with a larger variety of states, actions, and design goals.

Building on this work, several researchers have explored genetic programming for behavior trees in 2D games with simple objectives, such as Mario~\cite{colledanchise_learning_2018,zhang_behavior_2018} and Pac-Man~\cite{mcclarron_effect_2016,dockhorn_combining_2017}, a simple 2D shooter~\cite{estgren_behaviour_2017}, and -- with mixed results -- for the strategy games Zero-K~\cite{hoff_evolving_2016} and a simplified Battle for Wesnoth~\cite{oakes_practical_2013}. Colledanchise, Parasuraman, and Ögren developed a hybrid greedy search and genetic programming-based approach to create an agent for Super Mario~\cite{colledanchise_learning_2018}. They used genetic programming only to develop sub-trees when greedy search for a single action could not improve fitness. Zhang et al. investigated genetic programming with a modified crossover mutation operator, finding it effective on the Mario benchmark~\cite{zhang_behavior_2018} and, with the addition of constraints on the structure (similar to and-or trees~\cite{perez_evolving_2011}), on Pac-Man~\cite{zhang_learning_2018}. Outside of games, researchers have also tested genetic programming for evolving behavior trees to control robots~\cite{scheper_behavior_2016,jones_two_2018,iovino_learning_2021}, finding that they can create robust behavior despite uncertainty in action results. These efforts inspire our work to create a tool for evolving behavior trees for varied 3D games in Unreal Engine.

Perhaps the most similar work to ours is by Paduraru and Paduraru~\cite{paduraru_automatic_2019}, who created a framework to evolve behavior trees for 3D games for automated difficulty adjustment. They showed that their framework can develop varied trees, and they used human-in-the-loop testing to filter the resulting trees. They did not, however, aim to create a tool for designer-driven, iterative co-creation -- focusing instead on player-centric difficulty tuning -- and they did not evaluate their results with comparison to hand-designed trees.

Our work differs from these prior efforts in that its intent is not to evolve behavior trees to play a specific game as effectively as possible, but rather to enable co-creative, iterative, design-driven tool support for a flexible exploration of behavior. We aim to put designers in control of the fitness functions, the behavior tree nodes and building blocks, and whether and how the output of the tool is incorporated into the final AI. We also contribute an investigation of the experiential qualities of the resulting behavior, revealing insights on the challenges and opportunities for integrating this approach into modern game engine tools.

\section{The EvolvingBehavior Tool}

\begin{figure}
    \centering
    \includegraphics[width=\linewidth]{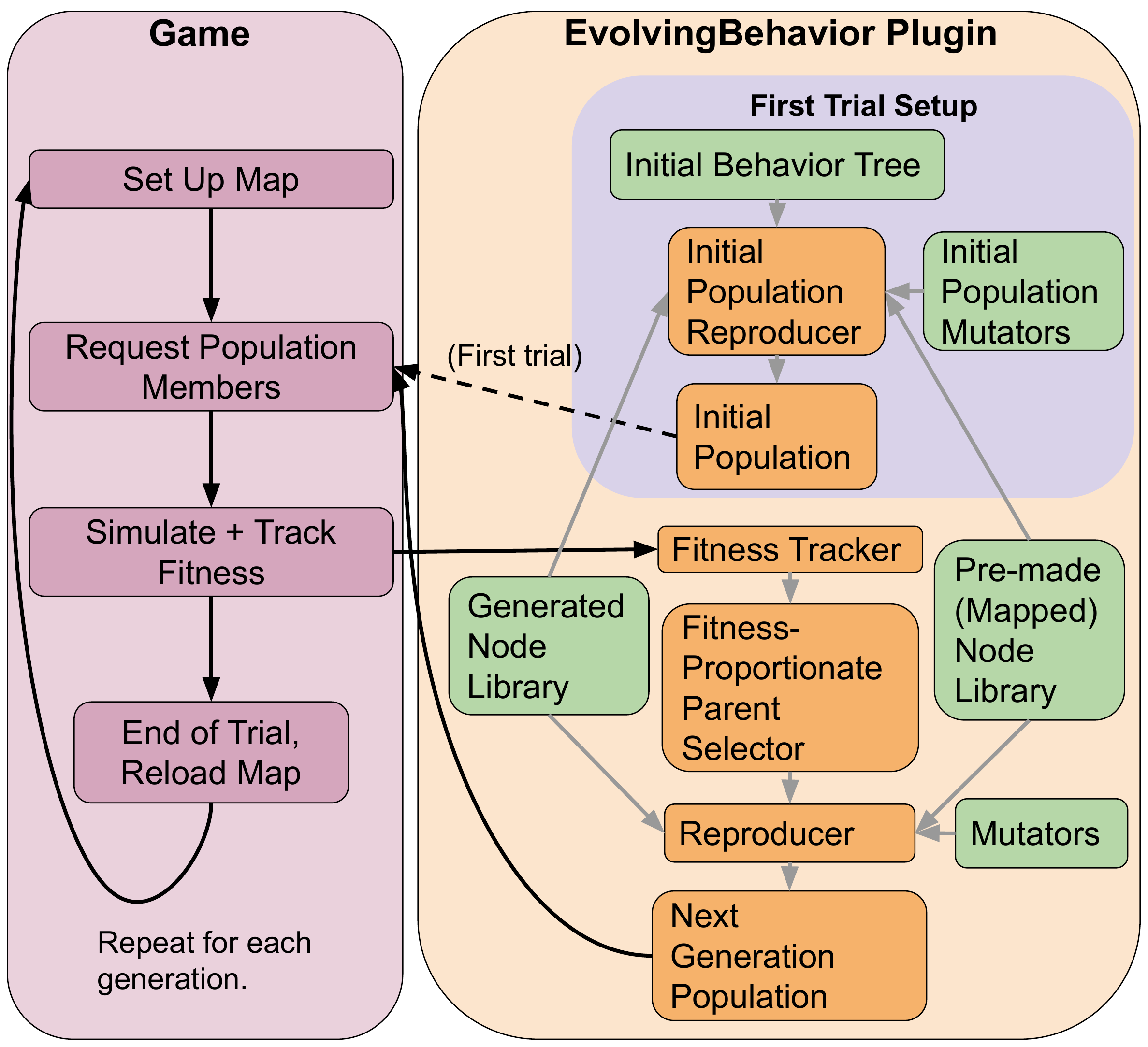}
    \caption{An overview of the process and data flow for evolving behavior trees using the EvolvingBehavior tool. Green nodes represent data configured by the designer, pink represent game-specific steps, and orange represent functionality and data provided by the EvolvingBehavior plugin. Grey arrows depict the flow of data within the plugin, whereas black arrows depict the flow related to the game.}
    \Description{Two large rectangles, one labeled ``Game'' and one labeled ``EvolvingBehavior Plugin,'' contain several smaller rectangles representing individual steps in the process of evolving behavior trees. These steps are described in more detail in the procedure section.}
    \label{fig:tooloverview}
\end{figure}

We created a tool (EvolvingBehavior) for evolving behavior trees, using genetic programming, in Unreal Engine 4. In EvolvingBehavior, a designer sets up and controls an experiment to evolve a behavior tree from a specific, hand-defined starting tree. An overview of this process is shown in Figure~\ref{fig:tooloverview}.

The designer controls the material with which the genetic programming algorithm will build modified trees by providing the initial tree, a library of additional pre-defined nodes, and templates for generating even more unique nodes with variable properties. The designer also controls other parameters and settings for mutation and the evolutionary process. These designer-controlled inputs are shown in green in Figure~\ref{fig:tooloverview}. Note that the initial tree need not be complete or functional -- it could even be empty -- as the tool will mutate and change it over time, seeking to improve its capabilities.

Though the designer must manage many parameters and decisions, we provide detailed documentation and plan to provide various default configurations to assist them. We also recognize the need for future work to test and improve the usability of the tool, which we discuss in more detail in Section~\ref{sec:futurework}. To enable a deep integration with the Unreal Engine user interface and allow for such future usability improvements, we chose to implement our own genetic programming system rather than to use an existing implementation. We now describe the technical implementation of the genetic programming process. 

\subsection{Representation}

Internally, we translate the built-in Unreal behavior trees into a compact and manipulable tree structure, a ``chromosome.'' This chromosome can flexibly store references to specific, pre-defined nodes (\textit{mapped nodes}) or to templates for generating nodes with properties that may range over various values (\textit{generated nodes}). We can then store, copy, and manipulate these chromosomes using the mutators to build a new generation.

The designer provides ``mapped'' nodes using Unreal's built-in behavior tree editor. The system translates these into chromosome form by simply tracking each as a unique identifier.

We also provide the ability to specify ``generated'' nodes, which are templates for creating a particular type of behavior tree node with random values for specific properties (variables). For each type of property -- integer, floating point number, boolean, and blackboard variable -- the tool provides an interface for specifying the range of possible values. (A ``blackboard,'' in this context, refers to a set of variables of various types that the behavior tree can access and modify to track its state for a specific agent.) When generating a new node, the tool stores in the chromosome a serializable representation of the names and associated values for that node's properties. These can then be translated back into properties of Unreal behavior tree nodes using reflection.

\subsection{Measuring Fitness, Flexibly}

Another area in which our system must provide designer control and flexibility is in measuring fitness. This is perhaps the most critical part of the system, the aspect that determines whether the results are helpful. After all, if the genetic programming is optimizing towards, say, ensuring that the agent survives, but the designer wants the agent to disregard its own safety in favor of attacking the player, the resulting behavior trees will be ill-suited for their task. Thus, no single fitness measure will suit every game, and each game must define its own.

Rather than prescribing a set fitness function, we provide developers with an event-driven architecture for defining their own. They define a key for each desired fitness measurement. Then, they send events to update the value of that fitness measure for each agent. At the end of the simulation, we compute a linear combination of the fitness measures with designer-defined weights. Moreover, this linear combination is only the default method of measuring fitness; it can be replaced, if necessary, to suit the needs of a particular game.

One important aspect of fitness that most games will share is a negative modifier for the complexity of the tree. Using tree size as a fitness penalty is one of the simplest ways to control ``bloat:'' growth of unnecessarily complex, often unused structures in the genetic program over many generations~\cite{poli_elitism_2008}. Bloat makes the resulting output less human-readable, which is clearly a problem if our goal is to suggest improvements to hand-built systems. We therefore give designers easy access to tree complexity, measured by the number of nodes, as a component of fitness.

\subsection{Selecting Parents}
After measuring each agent's fitness, we must select parents for the next generation. Specifically, we should select high-fitness parents more often, a process known as fitness-proportionate selection. We use a well-known selection scheme called ``tournament selection.'' Tournament selection is simple to implement, and it works with any fitness values, including negative fitness~\cite{poli_field_2008}.

To perform tournament selection, we first choose k members of the population at random. Then, we select the one with the best fitness as the final parent. The tournament size, k (4, in our experiments), determines how strongly the tournament selects for higher-fitness members of the population, and thus how much diversity will be lost in the next generation~\cite{blickle_comparison_1996}. The combination of simplicity, tunability, and input range resilience makes tournament selection a strong choice for a behavior tree evolution system, because it puts designers in control of the system without restricting them to a particular range for fitness scores.

Though some genetic programming systems separate parent selection from pruning the population, we chose to combine these two steps by replacing the entire population for each new generation. We do, however, retain a few of the most promising agents from the prior generation by elitism, which means copying the top-n\% most fit individuals directly to the next generation, unmodified. We hope, in this way, to avoid losing information about the most successful individuals due to random mutation, and to help reduce bloat.

\subsection{Reproduction and Mutation}
Once we have selected two parents, we produce a new child for the next generation. We begin by copying the primary parent exactly. Next, we may perform crossover and/or one or more point mutations to modify the child. In many genetic programming systems, the system performs at most one type of mutation per child, chosen probabilistically. Often, there is a significant probability of performing no mutation at all~\cite{poli_field_2008}. In our system, we instead chose to allow multiple mutation types to be applied, with a separate probability for each. We wanted to give designers more control over the mutation rates, without requiring them to worry about the precise total probability.

\subsubsection{Crossover} Crossover, in its most basic form, simply takes any random subtree from the child and replaces it with a random subtree from the second parent. Unlike in genetic algorithms with linear genetic codes, it is difficult to find a corresponding point on both individuals, so we do not require the subtrees to be related in position, or even size~\cite{poli_field_2008}. Koza suggests that, because there are many more leaf nodes than internal nodes, there should be a bias towards selecting internal nodes as crossover points. This will increase the average amount of the tree that is swapped, causing a larger effect on behavior~\cite{koza_genetic_1992}. In this work, we chose to use a crossover operator that chooses a node from a specific tree depth, where the depth is selected with uniform probability. This has been argued to be more effective than Koza's original crossover strategy~\cite{ito_depth-dependent_1998}, though there may be even more effective options we can explore in the future for crossover with non-uniform probability over the various tree depths~\cite{xie_depth-control_2011}.

\subsubsection{Point Mutation}
Among the many possible mutation operators, we chose to implement several types of point mutation: replacing, adding, or deleting a randomly selected node. In traditional genetic programming, these are often applied in a walk through the entire tree, where each node is mutated with some small probability~\cite{poli_field_2008}. In the current version of our system, we instead apply a random point mutation to a single node in the tree, if this mutation type is probabilistically selected.

We also provide point mutation operators for individual properties of generated nodes. Currently, we provide Gaussian perturbation mutators for floats and integers, with designer-specified standard deviation as a percentage of the current value (10\% in our tests). For boolean and blackboard variable properties, the mutator randomly selects a new value from among the available options.

\section{Experiments}

We performed a test of EvolvingBehavior by integrating it with the ``Epic Survival Game'' sample game by Tom Looman~\cite{looman_epicsurvivalgameseries:_2017}, and using it to evolve AI functionality for the simple zombie-like enemies. In this sample game, the player controls a 3D character using an over-the-shoulder style standard for the 3rd-person shooter genre. The player must survive for as long as they can in a level populated with simple zombie-like enemies. The enemies, in the standard game, can patrol a pre-set route, notice the player through sound or vision, and chase the player. If they get near the player, they cause damage. If they do 100 damage, they cause the player to ``die'' - in our tests, we set the players to then re-spawn after 2 seconds in a new random location. The player can shoot the zombies, and if they do enough damage, the zombie will ``die.''

Outside the behavior tree, the zombies' code tracks the nearest detected player using the blackboard. Thus, the tree needs only to determine actions, not to perform perception or target selection.

\subsection{Modifications to EpicSurvivalGame}

We created a simple ``human character'' behavior tree to stand in for the player characters. Approximately every half a second, it randomly selects a point near the character to move to, preferring points further away from zombies, visible to the character but not to the zombies, and closer to the current heading of the character. In this way, the characters randomly wander and avoid enemies. We also added the concept of stamina, which allows the player stand-in characters to run more quickly for a brief time when first being chased, before needing to slow down and recover. This requires the zombies to chase persistently in order to catch players.

To test EvolvingBehavior, we modified two maps included in ``Epic Survival Game,'' one a small and walled-in space with many obstacles shaped like boxes and shipping containers (the ``small map''), the other a larger and more open ``outdoor'' setting (the ``large map''), with only a few large rocks and buildings, as well as some trees, as obstacles. We created a third ``medium map'' by duplicating the small map, copying the space and obstacles four times, and removing the walls between the copies.

\subsection{Experiment Setup and Parameters}
\label{sec:experimentsetup}

For the initial behavior tree, to simulate the process of improving on a partial tree created by a human designer, we intentionally removed functionality from the zombie behavior tree originally included with ``Epic Survival Game.'' Specifically, we removed all nodes that allowed the characters to patrol and to chase the player, leaving some of the structure of the tree intact, but not enough for the characters to move in any way.

We set the initial population to use 10 iterations of random modification, such that at least one point mutation would occur with approximately 40\% probability, and crossover with 40\% probability. We provided the same mutators for the normal reproduction for each following generation, but halved to 20\% probability (specifically, 1.84\% for each point mutator, creating approximately a 20\% chance of at least one of the 12 point mutators being selected: $(1-x)^{12} = 0.8 \implies x \approx 0.0184$). To determine these, we first tried a grid search over crossover and point mutation rates of 10\%, 20\%, 40\%, and 80\%, finding that 10\% created the most stable results. Additional experimentation, however, revealed that 20\% tended to generate improvements faster with
acceptable stability. We use an ``elitism'' rate of 12\%, determined by preliminary experiments to provide a reasonable tradeoff between preserving strong agents versus retaining population variety.

\begin{figure}
    \centering
    \includegraphics[width=\linewidth]{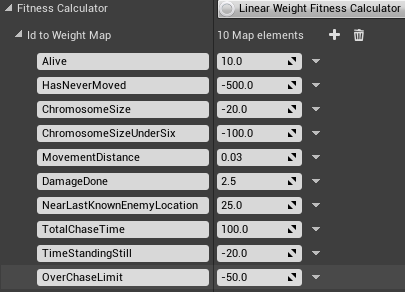}
    \caption{The identifiers and weight settings for the piecewise linear fitness function used in our experiments.}
    \Description{A set of text boxes list the names and associated weights of fitness function values for EvolvingBehavior.}
    \label{fig:fitness}
\end{figure}

We set fitness functions as shown in Figure~\ref{fig:fitness}. These piece-wise weighted linear fitness functions were determined by preliminary experimentation to reward zombies for moving and patrolling around the map, for chasing and doing damage to humans, and for searching near the last-known target location if they lose track of them. The fitness functions were also designed to penalize the zombies for failing to move, and for rapidly losing and re-starting a chase too many times (we detected the start and ending of chases based on consistent movement at an angle towards and within a short distance of a player, or the lack thereof). This latter penalty was based on research that found that designers would intentionally prevent NPCs from repeatedly chasing a player too frequently~\cite{partlan_design-driven_2021}. We also added a penalty for too complex or too simple a tree, to help reduce both bloat and the likelihood of trivial trees that would take too long to re-build into more complex behavior. The full calculations of these functions can be found in supplementary material\footnote{Archived at \url{http://hdl.handle.net/2047/D20450406}, or the up-to-date code repository can be accessed through \url{https://evolvingbehavior.npc.codes}}.

Next, we evolved agents using the above parameters, on each of the three maps, each with three different numbers of human characters (3, 6, and 12), for a total of nine experimental conditions. We set the number of zombies in the population as follows: Small had 20 zombies, Medium had 50, and Large had 40. These conditions were chosen to provide a variety of densities and conditions in which to evolve zombie behavior. Note that we used the same fitness function across all experimental conditions, whereas a game designer might choose in practice to tailor the fitness function more specifically to the map and number of agents involved -- we discuss the results and importance of these choices in Section \ref{sec:discussion}.

We provided a variety of behavior tree nodes, both mapped and generated, for the evolution to use -- nodes from the original zombie behavior tree from ``Epic Survival Game,'' such as nodes that would select random patrol points and move to them using Unreal's built-in pathfinding logic; as well as variations on other built-in Unreal behavior tree nodes, such as wait timers, decorators and actions for rotation or angle between objects, and movement directly towards a specified object or location (rather than full pathfinding). We intentionally included some less-effective nodes in order to test the evolution's ability to improve on partially-working behavior. To support the additional fitness rewards for searching near the last known location of a player after losing track of them, we also provided nodes to find a location near that point, and to intentionally forget that location. In total, we provided 8 mapped and 4 generated decorators, 13 mapped and 5 generated tasks, and the standard 2 types of composite nodes (selectors and sequences). We did not use service nodes in this experiment, though they are supported by the tool. The full set of nodes and detailed parameter settings can be found in supplementary material.

We chose to run evolution for 1000 generations, twice the number in the initial grid search, because this was generally past the point where fitness improvement plateaued in preliminary testing. We used 15x game speed and 90 seconds of simulated in-game time (6 seconds of real time), which was the same as the grid search and provided a reasonable trade-off between computation time and fidelity of gameplay. We chose to select a single resulting behavior tree from the many options created during evolution by finding the generation with highest average fitness, and selecting the highest-fitness behavior tree from that generation. We chose this strategy based on its general success in finding high-performing trees in our preliminary experimentation, but other strategies would certainly be possible -- for instance, selecting the highest-fitness tree from the last generation. Future research on co-creative tools could explore various strategies, such as enabling the designer to test various high-fitness trees to find one that meets their design goals.

We performed three runs of the evolution with different initial conditions, and averaged the results of the fitness comparisons between the three resulting trees for each condition. We chose to use this small number of runs due to the significant time required for each, and to limit the number of resulting trees so that we could manually inspect and qualitatively evaluate them over all nine experimental conditions. We believe this is also consistent with how a designer might use the tool -- they would not likely choose to re-run the evolution process many times, as this would lengthen iteration times and increase the complexity of sorting through the resulting behavior trees. We further discuss the limitations of this experimental design in Section~\ref{sec:limitations} below.

\subsection{Comparison With Manual Design and Random Baseline}

For comparison, three researchers created behavior trees manually, attempting to create behavior to meet the design goals as expressed by the fitness function. One researcher (R1) had experience creating game AI as a programmer at a game studio; the other two (R2 and R3) had no significant prior experience creating behavior trees. R1 briefly explained to R2 and R3 the general concepts and specific operation of behavior trees in Unreal Engine.

We also created behavior trees using a ``random baseline'' algorithm, which was the same as the standard evolution, except that parent selection was replaced with reproducing and mutating all trees from the previous generation, without any regard for fitness. All other settings were the same as for the evolved trees, as was the selection procedure to choose the final ``best'' tree from each run. Thus, the random baseline starts with and acts on the partial initial trees in the same manner as the evolutionary algorithm, but without the direction provided by a fitness function, enabling direct comparison of the results of the two algorithms.

We compared the fitness for the trees in each experimental condition, comparing the manual behavior trees, and the random baselines and evolved trees from the specific condition. We created the population of zombies using only the specified tree, and ran 100 trials without any changes to the trees (each trial being similar to fitness testing for a single generation in the original evolution), then averaged the results over all individuals and trials. We chose to run these fitness tests at 10x game speed to allow more computation time and thus provide a more accurate estimate of the fitness.

The code is written in C++ and Unreal Engine Blueprints, and compiled with the Unreal Build System. To evolve the trees, we used a single computer with an Intel i7 4810MQ procesor, 32 GB of RAM, and an NVidia GeForce 980M graphics card, running Windows 10. For the fitness tests, we used a separate computer with an Intel i7 8700K processor, 32 GB of RAM, and an NVidia GeForce 2070 graphics card, running Linux. Because the fitness testing was independent from the evolution, performance differences between the two computers have no impact on the results. Each evolutionary process of 1000 generations completed in approximately 2-3 hours, and fitness tests within 20 minutes.

\subsection{Qualitative Evaluation Method}
\label{sec:qualitativemethods}

This experiment has a relatively small sample size compared to many evolutionary algorithm experiments. Moreover, the goal is to show that evolution can produce similarly-capable behavior trees to human designers, rather than superior trees. Finally, we can only quantitatively measure the performance on a defined fitness function, which may not precisely match the human perception of behavior -- creating a fitness function that precisely encapsulates design goals is known to be a difficult problem~\cite{dill_what_2013,lehman_surprising_2020}. This all combines to challenge the relevance of standard statistical comparison techniques for evolutionary computation for this work. Thus, we chose to also perform a qualitative evaluation.

For our qualitative evaluation, two researchers (R1 and R4, one of whom had created a behavior tree for manual comparison, and one who had not) observed the zombies using the manual, random baseline, and evolved behavior trees in action against the human character AI, in the same scenarios in which the fitness comparisons were performed, but at normal game speed. We chose three conditions for this evaluation, such that each map and each density of human characters was represented. The researchers also played against the zombies directly. They wrote journals about their experiences and observations during and after these sessions.

Finally, R1 manually examined the behavior tree code, attempting to understand the reasons for the behavior. The two researchers then discussed their observations, and collaboratively summarized the results. We recognize the limitations of researchers acting as the game designers for this study and the need for future study with external designers, as discussed below in Section~\ref{sec:limitations}.

\section{Results}

\subsection{Quantitative Fitness Comparisons}

\begin{figure*}
    \centering
    \includegraphics[width=\linewidth]{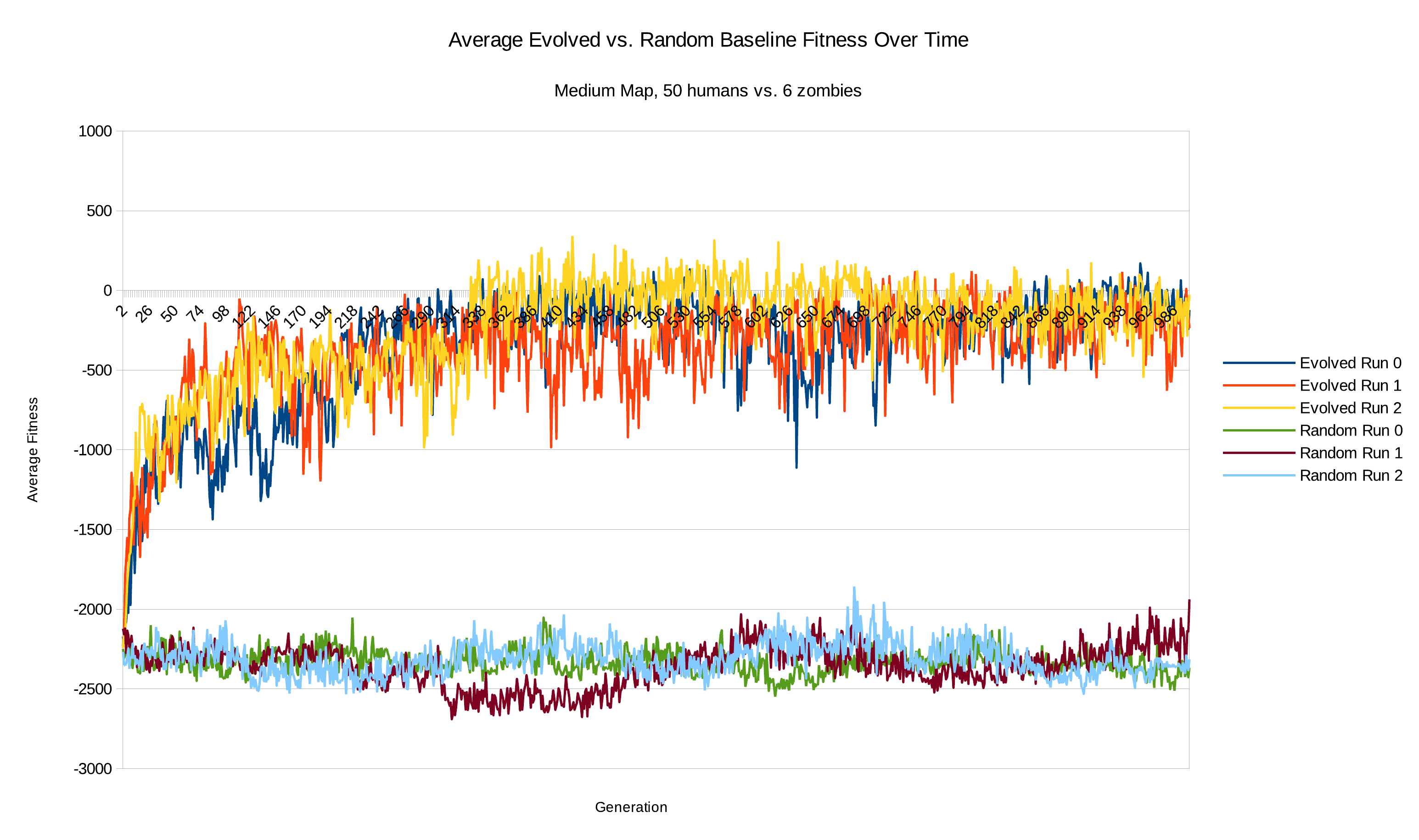}
    \caption{Graph of the average fitness per generation of the evolved vs. the random baseline zombies, on the Medium Map, with 50 zombies and 6 humans.}
    \Description{A line graph shows 6 different lines, representing average fitness of each generation of zombies in a particular experiment run, over 1000 generations. Each line is very jagged, with 3 of them representing the average fitness of Evolved zombies over time. These three lines start around -2000 fitness, then generally curve upward, and level out about mid-way through the 1000 generations shown, fluctuating around -750 to 250 fitness. The other three lines represent the Random Baseline zombies, and they also start around -2000 fitness, but generally stay and fluctuate around -2500 to -2000 fitness for the entire 1000 generations (never moving significantly upward).}
    \label{fig:fitness-overtime}
\end{figure*}

In Figure \ref{fig:fitness-overtime}, graphs show the average fitness over time for all runs of the Evolved and Random Baseline on the Medium Map, with 50 zombies and 6 humans, experiment setting. Due to space limitations, we provide the full fitness data from all runs in supplementary material\footnote{At \url{http://hdl.handle.net/2047/D20450406}}. The average fitness over time ceases to show further improvement well before the end of the 1000 generations.

\begin{figure}
    \centering
    \includegraphics[width=\linewidth]{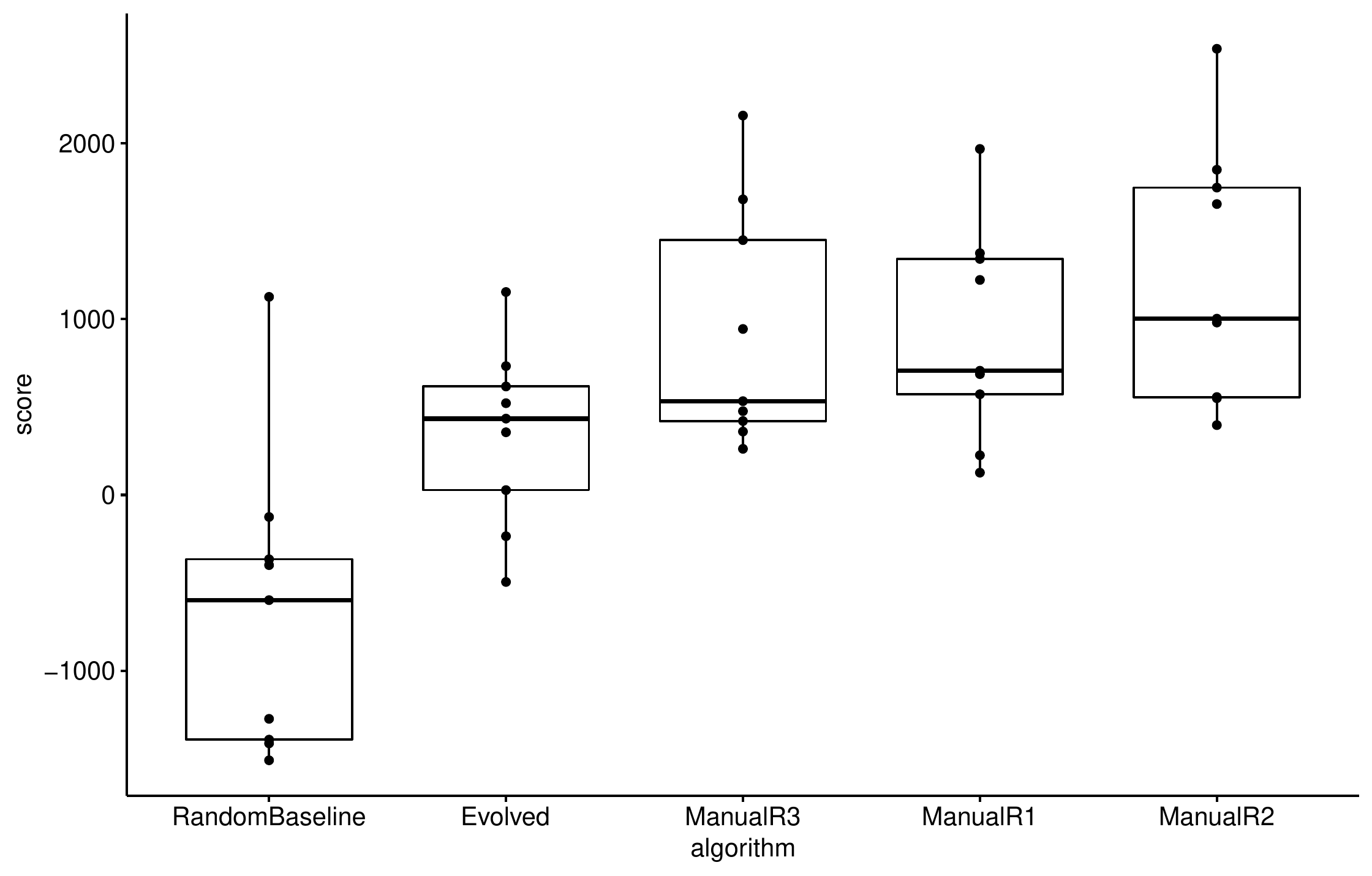}
    \caption{Box plots of the average fitness results for each algorithm across all problem configurations.}
    \Description{Five box plots are shown. The x-axis is labeled "algorithms," and shows the five algorithm names. The y-axis is labeled "score." The three Manual box plots have higher medians and show less variance. The median for the RandomBaseline box plot is lower and the plot shows more variance, and the median for the Evolved box plot is higher than the RandomBaseline, but lower than the Manual box plots, and shows the most variance.}
    \label{fig:fitness-boxplots}
\end{figure}

We performed a quantitative analysis to compare the fitness results for each algorithm. Descriptive statistics for the algorithms are shown as box plots in Figure \ref{fig:fitness-boxplots}. Because evolutionary algorithms do not generally result in scores that are normally distributed, we chose to use non-parametric statistics as recommended by the literature~\cite{derrac_practical_2011}, specifically the Friedman signed-ranks test, which found significant differences among the algorithms ($F_r=28, df=4, p < .0001$), with a large effect size (Kendall's W = 0.778). Therefore, we performed an all-pairs post-hoc test to determine which pairs of algorithms showed statistically-significant separation in fitness (because of its precise p-value calculation, we chose to use the test by Eisinga et al.~\cite{eisinga_exact_2017}). To adjust the post-hoc test for multiple comparisons, we used a Hommel correction (as discussed by Derrac et al.~\cite{derrac_practical_2011}). The results of the post-hoc test are shown in Table \ref{tbl:posthoc}.

Some statistically significant pairwise differences are shown by the post-hoc tests: between the Random Baseline and ManualR1 (p < 0.05), ManualR2 (p < $1e-07$), and ManualR3 (p < 0.01), and between ManualR2 and Evolved (p < 0.05). As a note, ManualR2 used a custom behavior tree node that was not available to the evolved behavior trees, as it combined selecting and moving to a random point nearby. We chose not to provide this node to the evolving behavior trees because it combined two functions (selecting a random location and moving there using full pathfinding) that would trivialize the implementation of random wandering behavior.

\begin{table}
\begin{tabular}{c|c|c|c|c}
.          &   Evolved  & ManualR1 &   ManualR2  & ManualR3 \\
ManualR1   &   0.4163   &  -       &   -         & -       \\  
ManualR2   &   0.0102*  & 0.2775   &   -         & -       \\
ManualR3   &   0.2775   & 0.7133   &   0.4163    & -       \\
Random     &   0.2775   & 0.0102*  &  5.2e-08*** & 0.0017** \\
\end{tabular}

\caption{Results of the all-pairs post-hoc test, showing the probabilities that the trees generated by each pair of algorithms have significantly different fitness. Asterisks indicate level of significance: * significant with p < 0.05, ** significant with p < 0.01, *** significant with p < 0.001}

\label{tbl:posthoc}
\end{table}

\subsection{Qualitative Evaluation}

\begin{figure*}
    \centering
    \begin{subfigure}{.3\linewidth}
        \includegraphics[width=\textwidth]{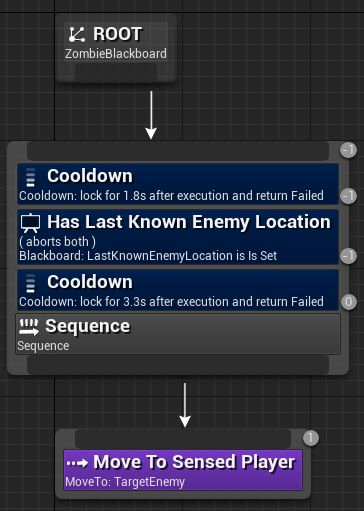}
        \caption{The Random0 tree for Medium50v3.}
        \label{fig:examplebts-Random0}
    \end{subfigure}
    \begin{subfigure}{.63\linewidth}
        \includegraphics[width=\textwidth]{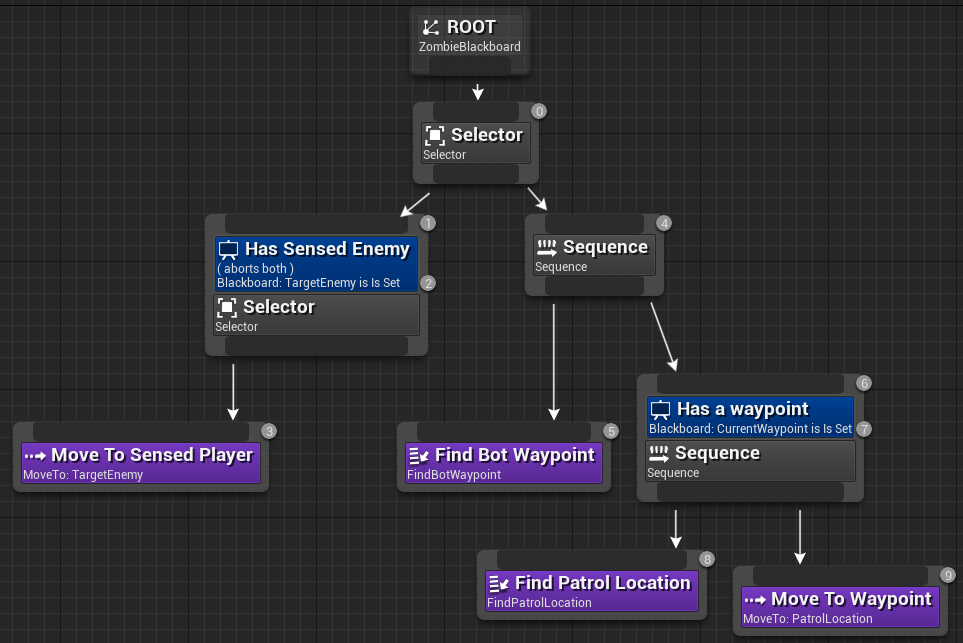}
        \caption{The ManualR3 behavior tree.}
        \label{fig:examplebts-ManualR3}
    \end{subfigure}
    \begin{subfigure}{.70\linewidth}
        \includegraphics[width=\textwidth]{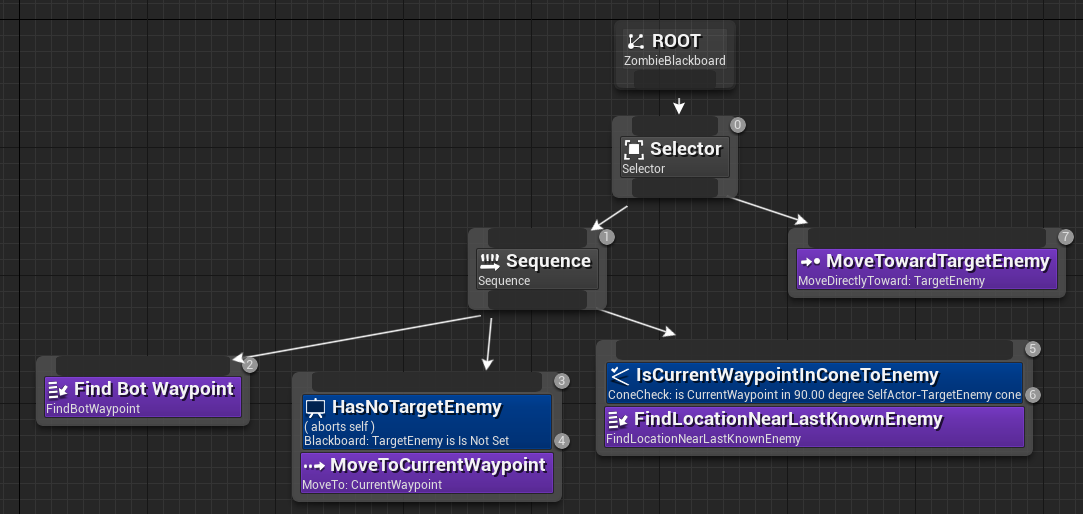}
        \caption{The Evolved2 tree for Medium50v3.}
        \label{fig:examplebts-Evolved2}
    \end{subfigure}
    \caption{Examples of behavior trees discussed in the qualitative evaluation: Random0 and Evolved2 from the Medium Map 50v3 condition, and the ManualR3 hand-designed tree.}
    \label{fig:examplebts}
    \Description{Three behavior trees are shown. Tree (a) is the Random0 behavior tree for the Medium50v3 condition. It contains a single task node ("Move to Sensed Player") under a Sequence node with several decorators. Tree (b) is the ManualR3 behavior tree. It contains a root selector, with the top priority being a "Move to sensed player" node under another selector, conditioned on "has sensed enemy." If that fails, it uses a sequence to "find bot waypoint," then, if "has a waypoint," to "find patrol location" and "move to waypoint" to that patrol location. Finally, tree (c) is the Evolved2 tree from the Medium50v3 condition. It has a root selector, with the first priority being a sequence of "find bot waypoint," followed by a "move to current waypoint" if "has no target enemy," and finally a "find location near last known enemy" if "is current waypoint in cone to enemy." Then, at lower priority in the root selector, a "move toward target enemy."}
\end{figure*}

Two researchers (R1 and R4) evaluated the behavior trees for three of the nine conditions (Medium Map 50v3, Large Map 40v6, and Small Map 20v12), selected for variety, as described in Section \ref{sec:qualitativemethods}. We present a summary of their findings, with relevant quotes. Figure \ref{fig:examplebts} shows some of the trees discussed below. Note that there are only three Manual behavior trees in total, but there are separate Evolved and Random trees for each condition. When we refer to, e.g., ``Evolved0,'' in Medium Map 50v3, we mean the evolved tree selected from the first run of that specific condition, and it is a different tree from, e.g., the ``Evolved0'' in Large Map 40v6.

In all three conditions, the three Manual trees exhibited mostly similar behavior. With all three, all zombies randomly walked around (sometimes walking, sometimes jogging) until they observed a player. When they spotted a human character, they immediately stopped wandering and chased that character. If they lost track of the human character, they resumed wandering. All were able to avoid static obstacles, though they sometimes ran into other zombies, especially in tight passages.

There were two noticeable differences between the specific behaviors of the three Manual behavior trees: in the distance and rapidity of wandering, and in their behavior when losing track of a player. ManualR1 zombies would ``stay near the place where they lost track, moving in a small radius and changing direction frequently (with a second's pause between each direction change), for 5-10 seconds'' (R1), after which ``they return to longer-distance random wandering'' (R1). ManualR2 and ManualR3 (Figure \ref{fig:examplebts-ManualR3}) zombies, on the other hand, would ``immediately turn around and move away (walking speed) in another direction'' (R1). ManualR1 and ManualR3 zombies tended to wander ``fairly long distances before changing direction'' (R1), whereas ManualR2 zombies would move a more variable distance before pausing and changing direction. ManualR1 and ManualR2 paused between direction changes when wandering (ManualR2 tended to pause longer); ManualR3 did not.

Inspection of the code showed that ManualR2's wandering behavior was controlled by a custom node, which used a random radius for selecting points to wander to, whereas the other trees used ``find waypoint'' and ``find patrol location'' nodes to find random hand-placed points on the map to move to, explaining the behavioral differences. Only ManualR1 used the ``find point near last known player location'' to wander near the last place a player was seen after losing track of them, which explains the difference in behavior in that instance.

\subsubsection{Medium Map 50v3}

In this condition, all Random Baseline zombies stood completely still until they observed a human character. In two cases, Random0 (Figure \ref{fig:examplebts-Random0}) and Random1, they would chase humans after noticing them in a manner fairly similar to the Manual trees (though Random1 would more easily get stuck on obstacles). Zombies using the Random2 tree were much less effective in chasing -- they would move towards a location relatively close to the human, but usually stop before reaching them. R1 wrote, ``If I made even a little effort to move away, they would fail to follow (but would slowly walk around a little where I left them),'' noting that they moved intermittently after losing track of a player. R4 wrote, ``There was very little patrolling behavior exhibited in the randomly generated tree and zombies seemed to only move in the presence of humans, but would remain stationary otherwise.'' Random0 and Random1 ceased moving as soon as they lost track of a human.

All three of the Evolved trees exhibited behavior fairly similar to the Manual trees: the zombies would wander randomly through the map, immediately turning to chase any humans they noticed. In two cases (Evolved1 and Evolved2, which is shown in Figure \ref{fig:examplebts-Evolved2}), the zombies would often run into obstacles while chasing: R1 wrote, ``I could lose zombies by getting them stuck on obstacles.'' Evolved0 did not have this problem, and were very similar to ManualR3 in their behavior. R1 noted a difference from ManualR1's behavior, in that after losing track of the human ``they did not stick around as long - seemed to lose interest immediately after.'' R4 agreed that the Evolved trees were similar to ManualR3, saying, ``The behaviors between the evolved tree and this [ManualR3] tree were very similar'', the only difference being that ``there is no fixed patrol area for [the Evolved] zombies, it seems as though in their patrol state, they simply move from spot to spot instead of revisiting an area that has been checked already.'' This latter difference might be explained by the presence of nodes to "find patrol location" and "move to" that location in the ManualR3 code, which adds a random offset to the manually-placed waypoints, whereas the Evolved trees moved to the waypoints directly.

Inspection of the code revealed that all three trees had evolved structures that effectively switched between chasing (as the top priority) and moving to a random waypoint, though Evolved1 and Evolved2 used the less effective ``move toward'' node to chase (which does not pathfind to avoid obstacles), and none used the ``last known enemy location'' to patrol near the last place they saw a human.

\subsubsection{Large Map 40v6}

As in Medium Map 50v3, all Random Baseline zombies only moved when chasing a human. R4 wrote, ``Similar to the observations in the medium map, the evolved trees led to zombies constantly patrolling areas in search of humans, while the random baseline trees again would chase enemies that made noise around them.'' The Random zombies were less effective in chasing than the Manual ones: ``[Random0] Zombies sometimes get stuck behind an obstacle while chasing, just running into it'' (R1). Random1 and Random2 also appeared to lose track of humans more easily, even if the human just changed angles abruptly. R1 wrote, ``When chasing, [Random1 zombies] appear to stop/pause slightly before reaching the human character, or go to where the character recently was rather than chasing to where it is now.'' Inspection of the code revealed that Random1 and Random2 used a node to move to the ``last known enemy location'' as their primary mode of chase, rather than moving to the currently observed enemy position, and Random0 used the ``move toward'' (non-pathfinding) node type.

In two of the three cases (Evolved0 and Evolved2), the Evolved trees were clearly more effective than the Random trees, exhibiting fairly similar behavior to the ManualR1 and ManualR3 trees. Evolved1, however, only patrolled randomly, with no chasing: ``The zombies totally ignore me'' (R1). Evolved0 was more likely to get stuck on obstacles, but Evolved2 was not: ``It's very hard to lose them, even if I go around a corner or obstacle, they often follow me effectively [...] They feel very similar to ManualR3 to play against'' (R1). R4 noted that the Evolved2 zombies ``tended to keep moving forward in a straight line until either they find a new player to chase or they end up near the very outskirts of the map,'' a difference from ManualR2, which changed directions more often.

R4 noticed that the evolved trees tended to lose track of the human if they crouched. This was notably different from the random trees, which would simply stop moving and continue attacking if the human was still nearby -- but the detection logic and the logic for dealing damage (which just checks if they are close to the human) are both handled outside the behavior tree entirely. We further discuss this question of division of responsibility between behavior trees and other parts of the NPC logic in Section~\ref{sec:discussion}. Inspection of the code showed that Evolved0 used a ``move toward'' node for chasing, whereas Evolved2 used a ``move to,'' and Evolved1 only used nodes that find and move to random waypoints.

\subsubsection{Small Map 20v12}

In this condition, none of the Random nor Evolved trees exhibited any patrolling behavior. They all focused solely on chasing. There were only a few minor differences in behavior: Random0 ``get stuck behind a barrier while chasing me sometimes'' whereas Random1 and Random2 are better at avoiding obstacles. Inspecting the code, Random1 and Random2 used ``move to'' nodes (which perform pathfinding), whereas Random0 used a ``move toward'' (which does not). The Random trees had only 2-3 relevant nodes in their behavior trees (a selector or sequence, and chasing node), with 8, 11, and 12 nodes that appeared irrelevant to the behavior, respectively.

Evolved0 exhibited the only notable movement when not in sight of a human: they ``appear to move slowly around the last place they saw a player sometimes'' (R1). R1 wrote that Evolved1 ``get stuck behind obstacles fairly often,'' whereas Evolved2 ``feel better at avoiding obstacles and chasing me around corners.'' R4 also noticed that Evolved2 was more similar to the Random1 and Random2 in avoiding obstacles, and that the Evolved2 zombies would attempt to haltingly move around the obstacles even if they did get stuck. Inspecting the code, Evolved1 and Evolved2 were less wasteful, with 4 and 5 apparently irrelevant nodes respectively. Evolved0 had only 3 relevant nodes of 15 (10 irrelevant nodes were decorators on the root node that did not appear to affect the behavior). However, Evolved0 used a ``move to last known enemy location'' node, which caused its aforementioned unique movement behavior.

\section{Discussion}
\label{sec:discussion}

The quantitative results are not sufficient to understand the performance of the evolution by themselves. Though the tests did not find a significant difference between the Evolved trees and the Random Baseline, they also did not differentiate the Evolved trees and the human-designed ManualR1 and ManualR3 trees. This indicates that the non-parametric statistics (which rely only on the ordering of the results, not the magnitude of the differences) were not powerful enough to truly differentiate the algorithms with a sample size of 9 experimental conditions, so we explored the data and trees in more detail to determine whether there were, in fact, clear differences.

Inspection of the boxplots shows that the evolved trees had a clearly higher median score (434, IQR 590) than the random baseline (-598, IQR 1026). In fact, one IQR below the Evolved median is still higher than the Random Baseline median. This gives us some confidence from the summary results alone that there is a clear difference between the Evolved trees and the Random Baseline.

In 6 of the 9 conditions (Large Map 40v3 and 40v6; Medium Map 50v3, 50v6, 50v12; and Small Map 20v6), the Evolved trees have much higher fitness than the Random Baseline -- in them, the gap in average fitness between the Evolved and Random Baseline trees ranges between 559 and 1912 (mean 1464, std. dev. 541). This indicates that in these 6 conditions, the Evolved trees developed clearly better behavior in at least one of the runs, and often more.

There are no cases where the Random Baseline outperformed the Evolved trees. One of the conditions where the difference is less clear, Small Map 20v12, had the highest ratio of human AI to zombies, and fitness results for the Random Baseline were relatively close to those of the human-designed trees. Due to this small fitness difference and the stochasticity of the evaluation, it is likely that evolution had trouble differentiating behavior trees. This indicates that a single, static fitness function may not be sufficient for all conditions -- in practice, a designer would likely choose to tailor their fitness function to the specific map and scenario.

The other two conditions that did not show clear differences were Large Map 40v12 and Small Map 20v3. It is somewhat less clear why these did not produce clearly effective trees -- aside from the fitness function, another possibility is that population sizes were not large enough to maintain diversity, or the behavior settled in a local maximum. It is possible that more mutators might help -- perhaps a subtree-growth mutator, which creates an entire random subtree at once, rather than relying on individual nodes being added by point mutations (which might have no immediate effect by themselves, and thus their genes may fail to reproduce before their use can be discovered). We suggest other ideas in subsection~\ref{sec:futurework}.

In one condition (Medium Map 50v6), the Evolved behavior trees had higher fitness (avg. 521) than both the ManualR3 (262) and ManualR2 (397) trees. In one other condition (Medium Map 50v3), the Evolved trees (avg. 356) outperformed the ManualR1 tree (127). Though these differences are much smaller than those between the Evolved trees and the Random Baseline, so we cannot claim that they represent clearly better behavior, this does indicate that either the Evolved trees were sometimes equivalently successful to the human-designed trees, or that our fitness function did not capture our precise design goals sufficiently to differentiate them. To determine which, we turn to the qualitative evaluation.

Our qualitative evaluation supports the quantitative evidence that the Evolved trees were more effective than the Random Baseline -- and that the best Evolved trees were similar in many ways to the Manual trees, especially to ManualR3. It also illuminates particular behavioral differences (in both directions) that are relevant to the design goals. Specifically, in Medium Map 50v3, all three evolved trees exhibited intelligent transitioning between random patrolling and targeted chasing of players in a similar manner to the Manual trees, as did two of three in Large Map 40v6. None of the Random trees achieved such complex behavior, in any condition investigated -- they could only chase (or in the case of MediumMap 50v3's Random2, move slowly around the last seen player location). 

While the Evolved trees were not perfectly efficient (usually including a few nodes not clearly relevant to the behavior), they were all manually inspectable, using a manageable number of nodes. In practice, the best Evolved trees could be further pruned, tweaked, and iterated upon by a designer (by hand and/or by being used as a starting point for further evolution). Even in the less successful Small Map 20v12 condition, the Evolved trees were mostly more parsimonious than the Random Baseline trees.

In Small Map 20v12, neither the Evolved nor Random trees exhibited random patrolling, but this was the condition where there were so many humans to chase that patrolling was unnecessary for high fitness. We expect that, in practice, a designer would choose to use a different ratio of humans to zombies, or to adjust the fitness function appropriately, making this condition the least representative of real use of the tool. This does further support, however, the observation that defining an effective fitness function is important and potentially difficult for designers. In this and other Small Map conditions, the small population size may also have led to low chromosome diversity. Replicating maps to allow larger populations, or testing many individuals over time rather than spawning all members simultaneously, might improve diversity. Quality-diversity algorithms~\cite{mouret_illuminating_2015}, or other non-population-based methods, could also be explored in future work to ameliorate this.

This difficulty of fitness function definition is further supported by the fact that none of the most successful Evolved trees replicated ManualR1's patrolling near the last known player location. However, ManualR1 was less effective (by median fitness over all conditions) than ManualR2, and was not clearly more effective than ManualR3 -- and the latter two only used random patrolling. Therefore, we posit that our fitness function was not tuned effectively to encourage patrolling near the last known enemy location (despite our attempt to reward that behavior). If we, as researchers creating the tool, have trouble defining the fitness function to match our design goals, we expect designers may likely have similar trouble. We expect this to be an important avenue for future research.

\subsection{Implications for Co-Creative AI Design}

This study also illuminates specific implications for developing co-creative AI design tools and raises questions for future research. First, the qualitative evaluation provides some insight on potential methods for choosing behavior trees from the results of the evolution. We used the strategy of selecting, from each run, the highest-fitness behavior tree from the generation with the highest average fitness. This produced several trees which exhibited reasonably similar behavior to our manually-designed trees but implemented that behavior in several different ways.

As researchers, we were able to manually evaluate these trees, understand their differences in behavior, and compare to our design goals. By presenting several options for designers to test, modify, and iteratively evolve, we believe we can build a mixed-initiative AI tool that retains full designer control over the resulting behavior -- though further user testing is necessary to determine the most effective modes of interaction for game designers. Speed of iteration is one particular concern that will require future study, as simulating gameplay over many generations can be very time-consuming.

Additionally, some evolved trees had more extraneous nodes, some fewer -- comparing fitness relative to tree size might allow us to filter to the most parsimonious trees, or post-processing (perhaps using static analysis for unnecessary nodes, similar to~\cite{colledanchise_learning_2018}) might allow further improvement of the trees before presenting them to the designer. We could also explore alternative methods of selection, such as using tree difference algorithms to search for varied (but high-fitness) trees, manual testing of a variety of trees (as in~\cite{paduraru_automatic_2019}), or quality-diversity based approaches~\cite{mouret_illuminating_2015}.

This study further reinforces the known challenge of finding appropriate parameter settings for the evolutionary process -- Poli, Langdon, and McPhee note that, while genetic programming is often fairly robust, it does require tuning for each individual application~\cite{poli_field_2008}. These parameters, such as mutation rates, number and size of generations, tournament size for parent selection, elitism, and the like, were not always obvious. Mutation rates, for instance, were not necessarily correctly selected by a grid search -- we ended up doubling them for the final experiment after further testing, due to the initial grid search being biased towards stability over innovation. Co-creativity will likely further complicate this, but may also provide new opportunities for addressing these difficulties. Designers have expressed the need for ``sensible defaults'' and hiding of unnecessary complexity in prior research~\cite{partlan_design-driven_2021}, which indicates the need to provide various default parameter settings, contextualized with the designer's goals. For example, a co-creative AI design tool could provide settings with low mutation rates designed for ``fine-tuning,'' and suggest those when the designer indicates that they are already mostly satisfied with a behavior tree, whereas it would suggest higher mutation rates for initial exploratory work.

Beyond parameter settings, the study also shows the importance of selecting component parts (e.g. nodes and mutators). One significant difference between the behavior of the ManualR2 and Evolved trees was the manual tree's use of a custom node that combined selection of and movement towards a random nearby destination. By controlling the level of granularity of the behavior tree nodes and the parameters to be tuned inside them, and iterating on the set of available behaviors, designers may be able to further improve and customize their NPCs. This may require some specific AI design knowledge, however, which a co-creative tool may need to help the designer acquire. Selecting specific mutators may be another area in which this expertise will need to be taught -- and there are mutator types in the genetic programming literature, such as subtree-growth mutators~\cite{poli_field_2008}, that we have not yet explored.

Finally, there are some aspects of non-player character behavior that, depending on the game, may not be included in a behavior tree at all, or that could be better handled in other architectures. For example, in ``Epic Survival Game,'' the perception, tracking, and selection of a target to chase was handled entirely in code outside of the behavior tree, as was damaging the player. In other games, attacks might be triggered by task nodes in the behavior tree. High-level decision-making logic for choosing behavior may also be handled outside the behavior tree in some games, e.g.~\cite{neufeld_evolving_2019}. This may mean that co-creative AI design tools cannot rely purely on genetic programming to evolve behavior trees if they wish to assist the designer with all aspects of the NPC design process. It may be necessary to apply genetic programming to additional architectures such as finite state machines, or to add other techniques more appropriate to those other control structures and types of logic.

\subsection{Ethical Considerations}
\label{sec:ethics}

While studying tools for AI in games may seem relatively benign at first glance, there are several areas for ethical concern related to this work. First, behavior trees are also used in robotics and simulation applications, some of which have been applied in military contexts~\cite{iovino_survey_2020}. EvolvingBehavior, however, is expressly created for application only to games, and we have not investigated any ethical controls nor considerations concerning its use in any serious real-world context -- nor especially in life-or-death situations. As a first effort towards preventing such misuse, the published version of this tool includes a statement that it may not be used for military, intelligence agency, and law enforcement uses. We also recognize the need for more comprehensive discussion and action to prevent misapplying games research (as per Cook~\cite{cook_social_2021}).

Additionally, we propose AI tools that directly impact how game designers work. We must consider the context of that work: game developers are often over-worked, laid off, and otherwise abused -- especially developers of marginalized identities~\cite{vysotsky_sobering_2018,ruberg_queer_2020}. Our goal is to create a tool that does not replace game designers, nor reduce their agency. Intentions, however, are not enough -- we must design features in collaboration with designers, empowering them to decide how the tool should improve their work. We should ensure that our tools specifically require and support designer creativity, and without promoting surveillance of designers, nor misuse of their outputs or data. We should also join with game developers to collectively advocate for better, safer, and healthier working conditions, more stability and support, and the end of abusive practices.

Finally, we should be careful to separate the metaphor of evolutionary algorithms from real evolution in nature (i.e., parent selection is entirely artificial and arbitrary, genes and mutation are simplistic and not directly informed by natural mechanisms, etc.). In this work, we artificially select and modify AI agents based on a ``fitness'' function, using common metaphors from genetic programming. This must not, however, be confused to condone application of artificial selection to people, especially as evolutionary algorithms research has, in fact, occasionally used eugenics as a metaphor (e.g.~\cite{huang_harmonious_2018}). We do not have space here to fully explore this topic, but we call for future work to further clarify the separation between evolutionary algorithms and real evolution, and to push back against eugenicist metaphor in genetic algorithms research.

\subsection{Limitations}
\label{sec:limitations}

There are several limitations of this study. First, and most importantly, we as researchers acted as designers in this study. Though at least one researcher has significant game development experience, our primary day-to-day work is not as game designers. The researchers also have significant prior knowledge of the tool, its capabilities, and its limitations. This prior knowledge impacted our design goals and expectations for the trees. This was important for this preliminary study, as it allowed us to deeply interrogate the capabilities of the tool and underlying evolution technology. However, it leaves for future work a true user study with independent game designers to determine the tool's usability and capability of working with them in a co-creative design process. Of particular concern are the modes of interaction with the tool: understanding and tweaking parameters, defining and tuning fitness functions, setting up and iterating on test environments, etc. It will be especially important to include in future study designers who have less experience directly implementing game logic in code.

Additionally, our experiments comprised a relatively small sample size of three runs each of nine experimental conditions, all of which were partially related to each other (in that they shared goals and game design, and the medium map was modified from the small map). We therefore could not establish statistical significance of the difference in several treatments' average fitness function scores using non-parametric statistics. However, as noted in Section \ref{sec:experimentsetup}, these experimental design choices enabled the qualitative evaluation, which provides a more detailed analysis of the results.

The tested game itself was also relatively simple, and was designed as an example, not for widespread play. This limited the complexity of the behavior trees: the agents did not need to engage with some common mechanics in 3D action games, such as finding or using items, or being distracted or hampered by players. Even within this limited gameplay, our qualitative evaluation does reveal clear differences between algorithms and behavior trees. However, future work should expand the variety of game environments and design goals tested, to further support the utility of this tool and discover any limitations as behavior tree complexity increases.

\subsection{Future Work}
\label{sec:futurework}

We discussed the potential for exploring the addition of new types of mutators, such as subtree-growth mutators, to potentially improve the diversity of the population and avoid local maxima. Another option for avoiding local maxima might be quality-diversity algorithms, which could potentially keep more varied trees in the population~\cite{mouret_illuminating_2015}. These algorithms have been used for procedural content generation in games~\cite{khalifa_talakat_2018,alvarez_interactive_2020}, and might well be applicable to this problem as well.

Above, we discussed the importance and difficulty of defining effective fitness functions to match the design goals for the game. We believe this will be an important question for future research, requiring careful user interface and/or domain-specific language design, and potentially benefiting from further visualization support. For this problem specifically, and for the general utility of the tool more broadly, we believe it will be crucial to involve designers directly in the design of the tool through participatory design, co-design, and usability testing methods. Relatedly, we noted the need for future research on defaults and tradeoffs for various parameters, inputs and behavior tree design choices, and types of evolution.

We also call for future work that explores the combination of behavior tree evolution with other architectures, generative methods, and AI techniques, to provide flexibility and support to designers in defining perception, decision-making, and action for NPCs outside of behavior trees. There may also be opportunity for testing the generalizability of evolved behavior trees across contexts: between multiple maps or levels, or -- perhaps with additional evolution for fine-tuning -- even across game modes or similar games. It will be particularly important to investigate all of these in the context of solo or small-team game development, where designers may not have access to sufficient game AI engineering support.

\section{Conclusion}

In this work, we have described EvolvingBehavior, a novel tool for evolving behavior trees in Unreal Engine. Through experimentation, with researchers acting as game designers, we explored in detail the capability of this tool to modify and generate behavior trees that acted to meet specific game design goals. We found that, in appropriate conditions, the tool was capable of evolving NPC behavior that was noticeably similar to our hand-designed behavior, though with some differences and variability. By performing a detailed qualitative evaluation of these results, we explored the specific reasons for this variability, finding several important considerations for future work in co-creative game AI design tools. EvolvingBehavior represents a first step towards AI-enabled tools that empower game designers to achieve their NPC design goals.

\begin{acks}
Special thanks to Alex Grundwerg and Isha Srivastava for contributing code to an early prototype of the tool used in this work, and to Erica Kleinman, Muhammad Ali, and Sabbir Ahmad for collaboration on research related to this project. Many thanks to Tom Looman for allowing us to use the ``Epic Survival Game Series'' code, on which the testbed game code is built.
\end{acks}

\bibliographystyle{ACM-Reference-Format}
\bibliography{references}

\end{document}